\documentclass{article}
\pdfoutput=1

\usepackage[preprint]{neurips_2024}

\usepackage[utf8]{inputenc} 
\usepackage[T1]{fontenc}    
\usepackage{hyperref}       
\usepackage{url}            
\usepackage{booktabs}       
\usepackage{amsmath,amssymb,amsfonts}       
\usepackage{nicefrac}       
\usepackage{microtype}      
\usepackage{xcolor}         
\usepackage{multirow}
\usepackage{natbib}
\usepackage{graphicx}
\usepackage{array}
\setcitestyle{numbers,square,comma}

\newcommand{\better}[1]{\textcolor{magenta}{#1}}
\newcommand{\worse}[1]{\textcolor{blue}{#1}}

\title{FNP: Fourier Neural Processes for Arbitrary-Resolution Data Assimilation}

\author{%
  Kun Chen \thanks{Work done while being an intern at Shanghai Artificial Intelligence Laboratory.} \\
  Fudan University \\
  \texttt{kunchen22@m.fudan.edu.cn} \\
  \And
  Tao Chen $^\dagger$ \\
  Fudan University \\
  \texttt{eetchen@fudan.edu.cn} \\
  \And
  Peng Ye \\
  Fudan University \\
  \texttt{yepeng20@fudan.edu.cn} \\
  \And
  Hao Chen \\
  Shanghai Artificial Intelligence Laboratory \\
  \texttt{chenhao1@pjlab.org.cn} \\
  \And
  Kang Chen \\
  Shanghai Artificial Intelligence Laboratory \\
  \texttt{chenkang@pjlab.org.cn} \\
  \And
  Tao Han \\
  Shanghai Artificial Intelligence Laboratory \\
  \texttt{hantao.dispatch@pjlab.org.cn} \\
  \And
  Wanli Ouyang \\
  Shanghai Artificial Intelligence Laboratory \\
  \texttt{ouyangwanli@pjlab.org.cn} \\
  \And
  Lei Bai \thanks{Corresponding author.} \\
  Shanghai Artificial Intelligence Laboratory \\
  \texttt{bailei@pjlab.org.cn} \\
}

\begin{document}

\maketitle

\begin{abstract}

Data assimilation is a vital component in modern global medium-range weather forecasting systems to obtain the best estimation of the atmospheric state by combining the short-term forecast and observations. Recently, AI-based data assimilation approaches have attracted increasing attention for their significant advantages over traditional techniques in terms of computational consumption. However, existing AI-based data assimilation methods can only handle observations with a specific resolution, lacking the compatibility and generalization ability to assimilate observations with other resolutions. Considering that complex real-world observations often have different resolutions, we propose the \textit{\textbf{Fourier Neural Processes}} (FNP) for \textit{arbitrary-resolution data assimilation} in this paper. Leveraging the efficiency of the designed modules and flexible structure of neural processes, FNP achieves state-of-the-art results in assimilating observations with varying resolutions, and also exhibits increasing advantages over the counterparts as the resolution and the amount of observations increase. Moreover, our FNP trained on a fixed resolution can directly handle the assimilation of observations with out-of-distribution resolutions and the observational information reconstruction task without additional fine-tuning, demonstrating its excellent generalization ability across data resolutions as well as across tasks.

\end{abstract}

\section{Introduction}

Accurately estimating the true state of complex and chaotic Earth systems is an important and challenging task, which can contribute to a better understanding of nature and improve forecasting by reducing the error of initial conditions. The most accurate human knowledge of the Earth's state comes from observations, which are inherently limited in their scopes due to practical constraints. Data assimilation, based on limited observational information and short-term forecasts (referred to as the \textit{background}), serves as the primary approach for state estimation~\cite{lorenc1986analysis, gustafsson2018survey, le1986variational}. Traditional data assimilation methods employed in operational systems include Kalman filters based on minimum variance estimation and variational methods based on maximum likelihood estimation~\cite{asch2016data, carrassi2018data, rabier2003variational}. Taking 3D variational (3D-Var) data assimilation as an example, data assimilation is regarded as an optimization problem under given conditions, aiming to find the \textit{analysis} $x_a$ that minimizes the objective function $J(x)$. It can be formulated as
\begin{equation}
x_a = x^* = \mathop{\arg\min}\limits_{x} J(x)
\end{equation}
\begin{equation}
J(x) = \frac{1}{2} (x-x_b)^T B^{-1} (x-x_b) + \frac{1}{2} (y-Hx)^T R^{-1} (y-Hx)
\end{equation}
where $B$ and $R$ correspond to the error covariance matrix of the background $x_b$ and observation $y$, respectively, and $H$ is observation operator that maps state variables to observational space, aligning the background and observations with different modalities (for example, satellites do not directly observe state variables such as wind speed) and resolutions.

With the significant achievements of machine learning in medium-range weather prediction~\cite{pathak2022fourcastnet, bi2023accurate, lam2023learning, chen2023fengwu, chen2023fuxi, ben2024rise}, data assimilation has gained increasing attention as one of the core components in building end-to-end global weather forecasting systems. Compared to traditional methods, machine learning-based data assimilation models offer the potential for competitive results with significantly reduced resource consumption and execution time~\cite{chen2023towards, huang2024diffda, xu2024fuxi}, making it a promising research direction with practical applications. Chen et al.~\cite{chen2023towards} proposed a data assimilation model for weather variables based on the idea of gated masks, and combined it with FengWu~\cite{chen2023fengwu}, an advanced AI-based weather prediction model, to build the first end-to-end AI-based global weather forecasting system. Subsequently, data assimilation models integrated with other AI-based weather prediction models were propsed~\cite{huang2024diffda, xu2024fuxi}. These methods have demonstrated performance and efficiency improvements through various experiments, but \textit{all of them can only assimilate observations with the same resolution as the forecasting model}. Therefore, they need to interpolate the observations onto grids of corresponding resolution through pre-processing in advance, and the pre-trained models do not have the flexibility and out-of-domain generalization to assimilate observations with other resolutions. The pre-processing step implements part of the function of the observation operator $H$ to perform resolution alignment, and it will introduce additional errors inevitably, thereby affecting the performance and generality of data assimilation methods.

Neural processes~\cite{garnelo2018conditional, garnelo2018neural, kim2018attentive, gordon2019convolutional, petersen2021gp} offer a promising and universal data assimilation framework for addressing the aforementioned challenges. Neural processes are a series of conditional generative models that continuously model the distribution of functions and fields based on paired coordinate-value conditions, and generate values at arbitrary target locations based on coordinate indices. Their flexible features that allow for grid or off-grid data are well-suited for assimilating observational data with diverse forms, without requiring any prior interpolation or mapping~\cite{vaughan2022convolutional, andersson2023environmental, scholz2023sim2real}. \textit{In this context, data assimilation is defined as the process of generating the analysis given both background conditions and observational information conditions.} The network models the comprehensive functional representation based on the two conditional inputs and decodes it to obtain the posterior distribution of the target. Compared to deterministic data assimilation, the modeling of distribution by neural processes can provide uncertainty estimates and further be used for ensemble data assimilation~\cite{hamill2006ensemble}. Moreover, data assimilation task degrades to observational information reconstruction when the background condition is missing. These two tasks can be broadly categorized as conditional generation, enabling their straightforward integration into a unified framework for direct application through simple fine-tuning.

In this paper, we propose the Fourier Neural Processes (FNP) for data assimilation with arbitrary-resolution observations. FNP is flexible to adapt to varying resolutions and can be extended to any conditional generation task. Leveraging the efficiency of the designed modules and flexible structure of neural processes, \textit{FNP achieves state-of-the-art (SOTA) results in data assimilation experiments with different resolutions}, and demonstrates increasing advantages over other models as the resolution and amount of observational information increase. The visualization of the analysis showcases the promising performance of FNP in capturing high-frequency information. Importantly, the FNP trained at a fixed resolution can be directly applied to data assimilation with other resolutions and observational information reconstruction task without fine-tuning, highlighting its excellent out-of-domain generalization. Additionally, ablation study for different modules and experimental settings validate the effectiveness and robustness of our approach.

\section{Related Work}

\paragraph{Machine learning for data assimilation.}
There exist strong mathematical similarities between machine learning and data assimilation, enabling their integration within a unified Bayesian framework~\cite{geer2021learning, arcucci2021deep}. With their powerful nonlinear fitting capabilities and low computational cost, machine learning techniques can both enhance traditional data assimilation methods and provide alternative algorithms~\cite{cheng2023machine, buizza2022data, ham2022partial}. Convolutional neural networks (CNNs) and recurrent neural networks (RNNs) are often employed as surrogate models to replace computationally expensive components in data assimilation, such as tangent-linear and adjoint models in 4D variational (4D-Var) data assimilation~\cite{hatfield2021building}, localization functions in ensemble Kalman filters (EnKF)~\cite{wang2023convolutional}, and error covariance matrices~\cite{cheng2022observation, penny2022integrating}. Implicit neural representations (INRs)~\cite{li2024latent} and various autoencoders (AEs)~\cite{peyron2021latent, melinc2023neural, amendola2021data} can offer efficient order reduction frameworks for latent assimilation to address the challenges of high-dimensional data. More recently, algorithms based on diffusion models have also provided new solutions for data assimilation driven by the advancements and maturity of AIGC~\cite{huang2024diffda, rozet2024score, qu2024deep}. However, all these studies are aimed at assimilating fixed-resolution observations, and \textit{we are the first to focus on arbitrary-resolution data assimilation}.

\paragraph{Machine learning for observational information reconstruction.}
Observational information reconstruction is the process of recovering missing values and obtaining complete field information from limited sparse observations. Traditional reconstruction methods primarily rely on kriging interpolation and principal component analysis-based infilling~\cite{kadow2020artificial}. As Kadow et al.~\cite{kadow2020artificial} successfully applied the image inpainting techniques in computer vision to reconstruct the global temperature data, deep learning has been widely used in various reconstruction tasks~\cite{ma2023newly, wegmann2023artificial, egli2022reconstruction, wang2022deep}. We associate the reconstruction with the data assimilation task here, and \textit{the flexibility of our method allows the FNP pre-trained on data assimilation task to be directly applied to the observational information reconstruction without fine-tuning and achieve promising performance}.

\paragraph{Neural processes family and its application in geoscience.}
Neural processes combine the advantages of neural networks and Gaussian processes, and have demonstrated excellent performance in function regression, image completion and classification tasks~\cite{garnelo2018conditional, garnelo2018neural}. Attentive neural processes (ANP)~\cite{kim2018attentive} enable the network to learn location-relevant representations by introducing the attention mechanism, which improves the accuracy of predictions and broadens the scope of modeling. Convolutional conditional neural processes (ConvCNP)~\cite{gordon2019convolutional} model translation equivariance in the data, adding an important inductive bias to the model and enabling zero-shot generalization to out-of-domain tasks. Evidential conditional neural processes (ECNP)~\cite{pandey2023evidential} replace the standard Gaussian distribution with a hierarchical Bayesian structure through evidence learning to achieve the decomposition of epistemic-aleatoric uncertainty. Neural processes and their variants, leveraging their unique advantages, have been successfully applied in geoscience tasks such as climate downscaling~\cite{vaughan2022convolutional}, sensor placement~\cite{andersson2023environmental} and observational information reconstruction~\cite{scholz2023sim2real}, and have shown promising performance. Here \textit{we apply the neural processes to arbitrary-resolution data assimilation for the first time and achieve SOTA performance}, further demonstrating its huge application potential.

\section{Methods}

\subsection{Model Overview}

\begin{figure}
    \centering
    \includegraphics[width=\linewidth]{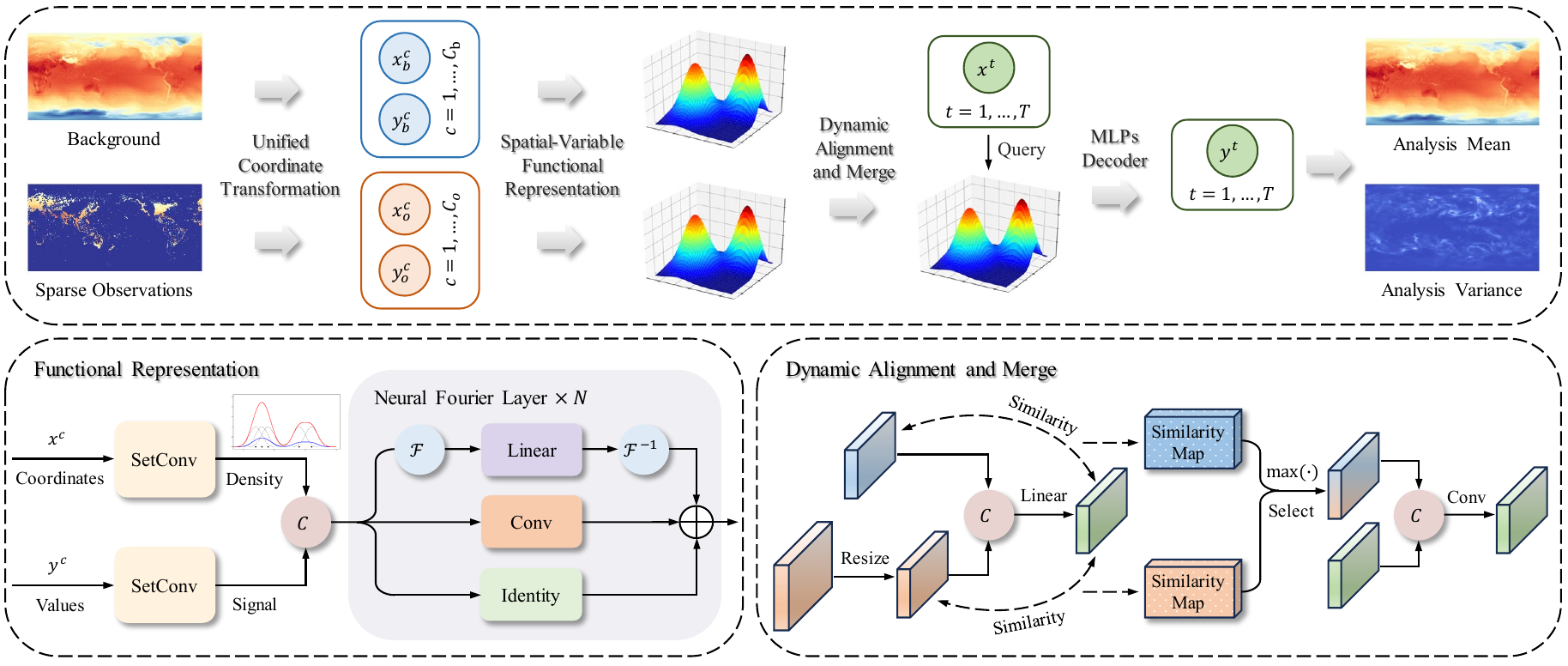}
    \caption{Overview of the network architecture of FNP. Unified coordinate transformation ensures spatial alignment of the background and observations, and extracts the coordinates and values of the conditional points. FNP models the two components of the conditional information globally to get their respective spatial-variable functional representation through \textit{SetConv} for data embedding and stacking of neural Fourier layers for deep feature extraction. The dynamic alignment and merge module integrates these functional representation based on similarity to shared features and aligns them into the target domain, resulting in a comprehensive functional representation over the target space. MLPs are finally employed to decode the functional representation and output the mean and variance of the analysis based on the coordinates of the target points.}
    \label{fig:network}
\end{figure}

The overall framework and model details of FNP are depicted in Figure~\ref{fig:network}. Initially, the background and observations undergo a unified coordinate transformation to obtain the coordinates $x^c$ and values $y^c$ of the conditional points when they are input into the network. This ensures spatial alignment of them, even in the presence of disparate resolutions, modalities, and data formats. Subsequently, FNP models the two components of the conditional information globally to get their respective spatial-variable functional representation. The dynamic alignment and merge (DAM) module integrates and aligns these functional representation into the target domain, resulting in a comprehensive functional representation over the target space. Finally, multi-layer perceptrons (MLPs) are employed to decode the functional representation and output the mean and variance of the analysis based on the coordinates $x^t$ of the target points. In the following subsections, we provide a detailed description of the process for modeling the functional representation and the internal structure of the DAM module.

\subsection{Spatial-Variable Functional Representation}

Modeling the functional representation involves two main steps: embedding the data sets into an infinite-dimensional function space and performing deep feature extraction. The former is accomplished through the \textit{SetConv} layer~\cite{gordon2019convolutional}, which is a generalized form of the standard convolutional layer extended to operate on sets. It takes a set of continuous coordinate-value pairs as input and outputs a function that can be queried at continuous positions. The \textit{SetConv} operation is permutation-invariant and includes an additional channel to estimate the density of the conditional points. When the input coordinates are discrete, \textit{SetConv} essentially degenerates into a standard convolutional layer, simplifying the model into the on-the-grid version~\cite{dubois2020npf}. We strongly recommend readers to refer to the theoretical proofs and derivations of ConvCNP~\cite{gordon2019convolutional}, and the project homepage of neural processes family~\cite{dubois2020npf} for a more detailed explanation.

The original deep feature extraction module is implemented using a standard CNN with residual structures. We choose to replace the basic convolutional layer with a more efficient neural Fourier layer (NFL) in FNP. Additionally, to address the multi-variable optimization problem in weather modeling tasks, we decouple the representations in spatial and variable dimensions to reduce the difficulty of network training. Below we provide further explanations on the motivations and implementation details of these design choices.

\paragraph{Spatial-variable decoupled representation.}
Data for different weather variables are usually stacked in the channel dimension, and direct data embedding will mix the spatial auto-correlation within variables with the inter-correlation among variables. An intuitive understanding is that explicitly separating the information in the spatial and variable dimensions allows for a clearer learning objective for each block, thereby reducing the difficulty of network training and fully unleashing the network's potential. In terms of implementation, we model a spatial functional representation separately for each meteorological variable, such as geopotential and temperature (surface variables are treated together as one variable), and model a variable functional representation that encompasses all variables, which are then concatenated together. The benefits of this approach have been confirmed in our experiments. We found that the spatial-variable decoupled (SVD) representation achieves better performance with fewer parameters and faster convergence speed. The detailed comparison of performance can be seen in Table~\ref{ablation study for modules}.

\paragraph{Neural Fourier layer.}
The smoothness of neural network outputs poses a challenging drawback in weather modeling tasks, and the background generated by AI-based forecasting models also tends to be smoother. In our experiments, we find that neural processes also struggle to overcome the issue of smoothness. To address the desire for high-frequency information, we choose to introduce the Fourier neural operator~\cite{li2020fourier}. Besides, operations in the frequency domain can also bring additional advantages in terms of global receptive fields for models based on CNN. Therefore, in addition to the convolutional operation, each neural Fourier layer consists of a branch for linear operation in the frequency domain and a branch for identity mapping to preserve high-frequency details as much as possible~\cite{he2016deep}.

\subsection{Dynamic Alignment and Merge}

The DAM module aligns functional representations from two conditional domains to the target domain for obtaining outputs at the target locations. In data assimilation tasks, the analysis typically shares the same resolution and modalities as the background, and the previous data embedding has already mapped inputs that may have different modalities into the same feature space. Therefore, it is only necessary to align the functional representation of the observation in the spatial resolution. We choose to use interpolation to adjust the spatial dimensions' size as it can accommodate inputs of arbitrary resolutions, thereby enhancing the dynamics and generality of the model. \textit{Interpolation in the feature space differs fundamentally from that in the original observational space because the former has already extracted helpful information and contains redundancy to support dimensionality reduction, while the latter compresses valid information and missing values to the same extent.} The performance of data assimilation with different resolutions in Table~\ref{arbitrary data assimilation} provides proof for this. As the amount of observational information increases, our model achieves significant improvements, while other models do not. A linear layer extracts shared features $\dot{y}$ from both parts after alignment, which are then used to calculate similarities with their respective feature components. The similarity calculation is performed in the channel dimension as the spatial distribution of information differs significantly between them, with the background having a more uniform spatial distribution while the observation exhibits greater spatial variability. In our implementation, the feature similarity is represented by the Euclidean distance between the two features, i.e.,
\begin{equation}
Sim_{h,w} = \sqrt{\sum_{i=1}^{k} (y_{h,w,i}-\dot{y}_{h,w,i})^2}
\end{equation}
where $h$ and $w$ denote the indices for each grid point along the longitudinal and latitudinal directions, respectively, and $k$ is the dimension of data embedding. The relative values of the similarity map then determines the selection of features. Specifically, features that are more similar to shared features will be retained, while features that are less similar will be discarded, that is,
\begin{equation}
y_{h,w}=\left\{
\begin{aligned}
y_{h,w}^b, & & Sim_{h,w}^b \ge Sim_{h,w}^o \\
y_{h,w}^o, & & Sim_{h,w}^b < Sim_{h,w}^o
\end{aligned}
\right.
\end{equation}
The dynamically filtered features will be spliced together with the shared features and sent to a convolutional layer for spatial smoothing, and the result will be used as the functional representation in the target domain for decoding and output.

\section{Experiments}
\label{section: experiments}

\subsection{Experimental Settings and Implementation}

\paragraph{Data preparation.}
We demonstrate the effectiveness of our methodology on the ERA5 dataset~\cite{hersbach2020era5}, a global atmospheric reanalysis archive containing hourly weather variables such as geopotential, temperature, wind speed, humidity, etc. We choose to conduct experiments on a total of 69 variables, including five upper-air variables with 13 pressure levels (i.e., 50hPa, 100hPa, 150hPa, 200hPa, 250hPa, 300hPa, 400hPa, 500hPa, 600hPa, 700hPa, 850hPa, 925hPa, and 1000hPa), and four surface variables. Specifically, the upper-air variables are geopotential (z), temperature (t), specific humidity (q), zonal component of wind (u) and meridional component of wind (v), whose 13 sub-variables at different vertical level are presented by abbreviating their short name and pressure levels (e.g., z500 denotes the geopotential at a pressure level of 500 hPa), and the surface variables are 10-meter zonal component of wind (u10), 10-meter meridional component of wind (v10), 2-meter temperature (t2m) and mean sea level pressure (msl). A subset of ERA5 dataset for 40 years, from 1979 to 2018, is chosen to train and evaluate the model.

\paragraph{Experimental settings.}
The advanced AI-based weather forecasting model, FengWu~\cite{chen2023fengwu}, is used as the surrogate model to generate the background. The observations are simulated by adding a proportional mask to ERA5, and the default setting corresponds to 24-hour forecast lead time and 10\% observations. In other words, the background used for data assimilation is produced by FengWu (with 6-hour interval) through four auto-regressive iterative predictions based on ERA5 data from one day ago. The observational space usually has higher spatial resolution than the state space in actual operation systems. Therefore, the resolution of the forecasting model and background is set to 1.40625° ($128\times256$ grid points) so that we can conduct experiments using observations with different resolutions such as 0.25° ($721\times1440$ grid points) to verify the assimilation performance with arbitrary resolution. 

\paragraph{Model training and evaluation.}
The FNP model is implemented based on the open-source code of the neural processes family project~\cite{dubois2020npf}, and trained for 20 epochs using the AdamW optimizer~\cite{loshchilov2018fixing} with a learning rate of 1e-4. We divide the ERA5 data from 1979-2015 as the training set, 2016-2017 as validation set, and 2018 as test set. The training is run on 4 NVIDIA Tesla A100 GPUs with a global batch size of 16, and takes approximately 2.5 days. The inference only needs a few minutes to perform data assimilation for a whole year on single A100 GPU. The dimension of data embedding for default setting is 128 and the number $N$ of NFLs is 4, and a Gaussian likelihood is used with a negative log-likelihood (NLL) loss. We evaluate the performance of models by calculating the overall mean square error (MSE), mean absolute error (MAE), and the latitude-weighted root mean square error (RMSE) which is a statistical metric widely used in geospatial analysis and atmospheric science~\cite{rasp2020weatherbench, rasp2023weatherbench}. Given the estimate $\hat{x}_{h,w,c}$ and its ground truth $x_{h,w,c}$ for the $c$-th channel, the RMSE is defined as
\begin{equation}
\resizebox{0.65\hsize}{!}{$
\operatorname{RMSE}(c) = \sqrt{\frac{1}{H\cdot W}\sum\nolimits_{h,w} H \frac{\operatorname{cos}(\alpha_{h,w})}{\sum_{h'=1}^{H} \operatorname{cos}(\alpha_{h',w})}(x_{h,w,c} - \hat{x}_{h,w,c})^{2}}
$}
\end{equation}
where $H$ and $W$ represent the number of grid points in the longitudinal and latitudinal directions, respectively, and $\alpha_{h,w}$ is the latitude of point $(h,w)$.

\subsection{Arbitrary-Resolution Data Assimilation}

We validate the performance of models by assimilating 10\% observations with resolutions of 1.40625°, 0.703125°, and 0.25°, respectively, onto a 24-hour forecast background with 1.40625° resolution. Table~\ref{arbitrary data assimilation} provides a quantitative comparison of the analysis errors between FNP and other models. The first row corresponds to the error level of the background. When assimilating observations with the same resolution as the forecasting model, \textit{FNP achieves SOTA results (indicated in bold) in terms of overall MSE, MAE, and RMSE metrics for the majority of variables}. Since Adas~\cite{chen2023towards} is not flexible enough to support inputs with different resolutions, we follow its common practice to interpolate the observations and average the observations falling within the corresponding grid range when assimilating higher-resolution observations. Despite this, Adas still produces results with significantly high errors, so we only present the performance after fine-tuning to the interpolated observations. This indicates that Adas lacks the ability of out-of-domain generalization. In contrast, FNP and ConvCNP~\cite{gordon2019convolutional}, with their flexible structures, can assimilate observations with different resolutions directly without interpolation. Therefore, the table presents the results for both cases with and without fine-tuning for FNP and ConvCNP.

\begin{table}[h]
  \caption{Quantitative performance comparison for arbitrary-resolution data assimilation. The best performance are shown in bold while the second best is underscored. Red color indicates the improved assimilation results compared to that with 1.40625° resolution, and blue color indicates worse results.}
  \label{arbitrary data assimilation}
  \centering
  \resizebox{\linewidth}{!}{
  \setlength{\extrarowheight}{3pt}
  \begin{tabular}{l|c|cc|cccccccc}
    \toprule
    \multirow{2}{*}{Model} & \multirow{2}{*}{Resolution} & \multirow{2}{*}{MSE} & \multirow{2}{*}{MAE} & \multicolumn{8}{c}{RMSE} \\
    & & & & z500 & t850 & t2m & u10 & v10 & u500 & v500 & q700 ($10^{-4}$) \\
    \midrule
    Background & 1.40625° & 0.0288 & 0.0861 & 45.455 & 0.7200 & 0.7790 & 0.9336 & 0.9645 & 1.7278 & 1.7535 & 6.7220 \\
    \midrule
    Adas~\cite{chen2023towards} & 1.40625° & 0.0221 & 0.0705 & \textbf{22.930} & \textbf{0.6323} & 0.7198 & 0.8198 & 0.8369 & 1.4673 & 1.4489 & 6.4955 \\
    ConvCNP~\cite{gordon2019convolutional} & 1.40625° & 0.0252 & 0.0795 & 31.253 & 0.6831 & 0.7662 & 0.8334 & 0.8553 & 1.5770 & 1.5876 & 6.5717 \\
    FNP (ours) & 1.40625° & \textbf{0.0202} & \textbf{0.0664} & 26.910 & 0.6600 & \textbf{0.7049} & \textbf{0.7558} & \textbf{0.7665} & \textbf{1.4263} & \textbf{1.4450} & \textbf{6.4691} \\
    \midrule
    Adas~\cite{chen2023towards} & 0.703125° & \better{\underline{0.0144}} & \better{\underline{0.0614}} & \worse{26.423} & \better{0.5981} & \better{\underline{0.6696}} & \better{0.6631} & \better{0.6552} & \better{\underline{1.2935}} & \better{\underline{1.1975}} & \better{\underline{5.3798} \underline{\small{$\downarrow$17.2\%}}} \\
    ConvCNP~\cite{gordon2019convolutional} w/o fine-tuning & 0.703125° & \worse{0.0378} & \worse{0.1134} & \worse{142.38} & \worse{0.9106} & \worse{0.8471} & \worse{1.2264} & \worse{1.1115} & \worse{2.9765} & \worse{2.6264} & \better{5.7900 \small{$\downarrow$11.9\%}} \\
    ConvCNP~\cite{gordon2019convolutional} & 0.703125° & \better{0.0202} & \better{0.0712} & \better{\underline{25.452}} & \better{\underline{0.5923}} & \worse{0.7861} & \better{\underline{0.5812}} & \better{\underline{0.6339}} & \better{1.3795} & \better{1.4331} & \better{5.7562 \small{$\downarrow$12.4\%}} \\
    FNP (ours) w/o fine-tuning & 0.703125° & \better{0.0150} & \better{0.0618} & \worse{52.075} & \better{0.6021} & \worse{0.7385} & \better{0.7052} & \better{0.6833} & \better{1.3723} & \better{1.3036} & \better{5.4333 \small{$\downarrow$16.0\%}} \\
    FNP (ours) & 0.703125° & \textbf{\better{0.0085}} & \textbf{\better{0.0419}} & \textbf{\better{20.710}} & \textbf{\better{0.5805}} & \textbf{\better{0.5310}} & \textbf{\better{0.4457}} & \textbf{\better{0.4406}} & \textbf{\better{0.9032}} & \textbf{\better{0.9180}} & \textbf{\better{4.5296 \small{$\downarrow$30.0\%}}} \\
    \midrule
    Adas~\cite{chen2023towards} & 0.25° & \worse{0.0231} & \worse{0.0766} & \worse{34.694} & \worse{0.6979} & \worse{0.7728} & \worse{\underline{0.8455}} & \worse{\underline{0.8537}} & \worse{\underline{1.5864}} & \worse{\underline{1.5265}} & \worse{6.5019 \small{$\uparrow$0.10\%}} \\
    ConvCNP~\cite{gordon2019convolutional} w/o fine-tuning & 0.25° & \worse{0.0651} & \worse{0.1514} & \worse{234.50} & \worse{1.3208} & \worse{0.8705} & \worse{1.7134} & \worse{1.5122} & \worse{4.1316} & \worse{3.6655} & \worse{6.6619 \small{$\uparrow$1.37\%}} \\
    ConvCNP~\cite{gordon2019convolutional} & 0.25° & \worse{0.0280} & \worse{0.0831} & \better{\underline{26.437}} & \better{\underline{0.6326}} & \better{0.7429} & \worse{0.9283} & \worse{0.9640} & \worse{1.7265} & \worse{1.7533} & \better{6.3612 \small{$\downarrow$3.20\%}} \\
    FNP (ours) w/o fine-tuning & 0.25° & \better{\underline{0.0196}} & \worse{\underline{0.0729}} & \worse{84.516} & \worse{0.6992} & \better{\underline{0.6912}} & \worse{0.9264} & \worse{0.8687} & \worse{1.8315} & \worse{1.6723} & \better{\underline{6.1761} \underline{\small{$\downarrow$4.53\%}}} \\
    FNP (ours) & 0.25° & \textbf{\better{0.0058}} & \textbf{\better{0.0339}} & \textbf{\better{19.996}} & \textbf{\better{0.4549}} & \textbf{\better{0.4999}} & \textbf{\better{0.3839}} & \textbf{\better{0.3753}} & \textbf{\better{0.6308}} & \textbf{\better{0.5320}} & \textbf{\better{2.9614 \small{$\downarrow$54.2\%}}} \\
    \bottomrule
  \end{tabular}
  }
\end{table}

In order to better understand and explain the performance of different models, we add different colors to represent the variations in results compared to that with 1.40625° resolution (blue indicating worse results, i.e., increased errors, and red indicating improved results). For the q700 variable, we additionally annotate the percentage of error increase or decrease. It is worth noting that with increasing resolution, the same ratio of observations implies a greater number of absolute observations and a larger amount of information. However, during the interpolation process, averaging the observation values within a region does not guarantee a reflection of the overall conditions unless there are observations at all points within the region. Therefore, interpolation inevitably leads to information loss, and the amount of lost information is negatively correlated with the number of observations within the region. Based on the balance between these two factors, the fine-tuned Adas exhibits different trends with two different resolutions: the results generally improve when assimilating observations with 0.703125° resolution, while all the results worsen when assimilating observations with 0.25° resolution. This indicates that as the resolution increases gradually, the impact of information loss due to interpolation becomes more significant and surpasses the positive effect of increased observation quantity. In practice, the observational data used in operational systems usually have resolutions of 0.1° or even higher, while the resolution of the most commonly used forecasting models is 0.25°. Therefore, the information loss caused by interpolation in existing methods is a very common and urgently addressed issue. 

In contrast, \textit{the fine-tuned FNP not only achieves SOTA performance in all metrics (including the z500 and t850 variables, in which FNP does not reach the optimum results with 1.40625° resolution), but also improves all assimilation results with the largest magnitude of error reduction}. The performance differences between FNP and other models increases significantly with increasing resolution, and the RMSE decreases on some variables such as v10, u500, v500 and q700 are even more than 50\% when assimilating observations with 0.25° resolution. Furthermore, as the absolute number of observations increases, providing more information, FNP is the only model that consistently improves the performance of data assimilation. This plays a crucial role in practical applications, as it means that all deployed observation instruments can be fully utilized, reducing the waste of human, material and financial resources as much as possible.

\begin{figure}
    \centering
    \includegraphics[width=\linewidth]{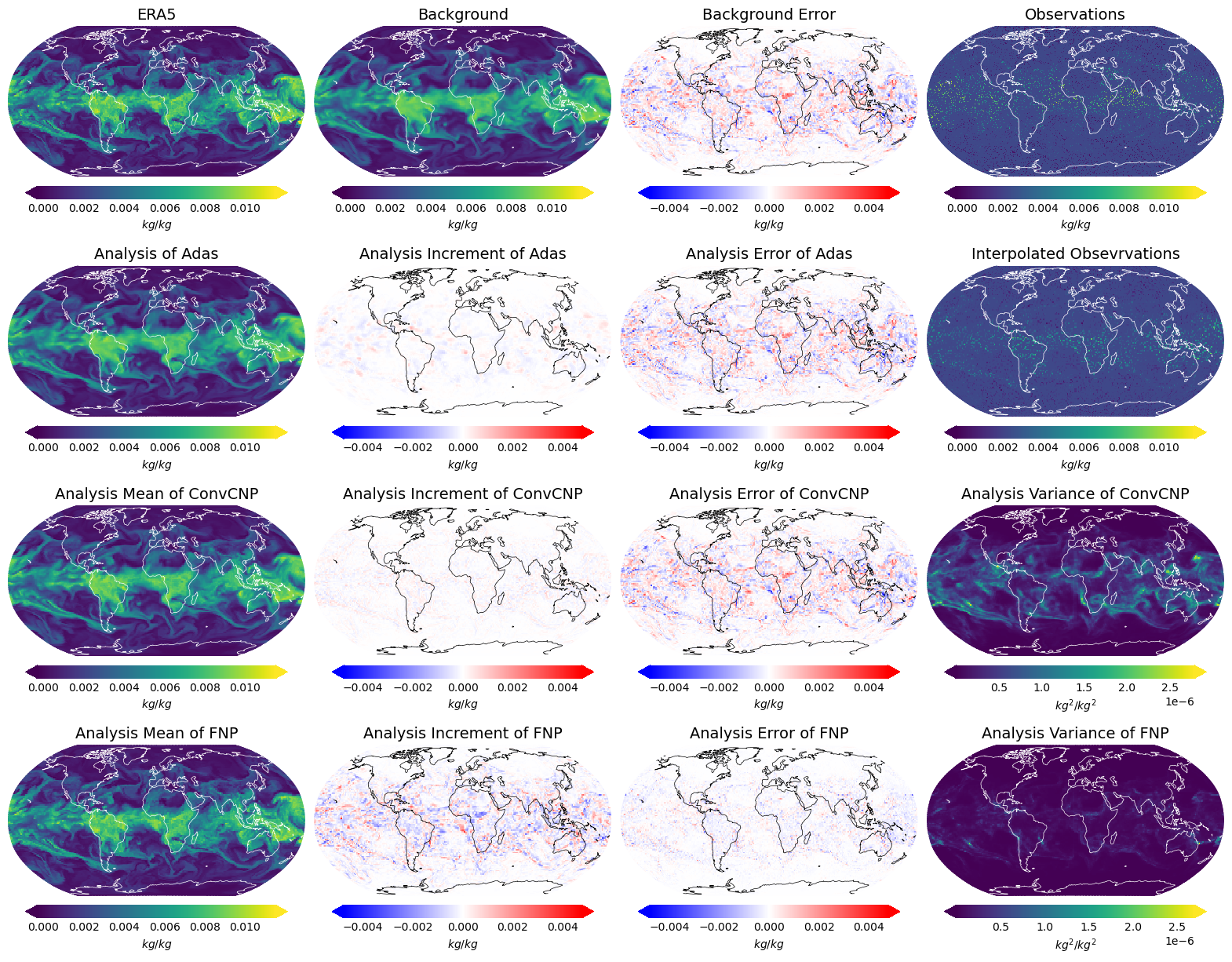}
    \caption{Visualization of assimilation results by different models for q700. The visualization date-time is randomly selected at 2018-04-02 06:00 UTC. The first row shows the ERA5 (ground truth), background, background error and observations with 0.25° resolution. Other rows show the assimilation results of different models.}
    \label{fig:visual_q700_721}
\end{figure}

The results without fine-tuning reflect the out-of-domain generalization capability of models, as they have not encountered observational data with other resolutions during training. As the resolution increases, the discrepancy between the samples used for testing and the visible samples during training becomes larger, leading to a gradual decline in performance. This regular pattern can be observed in the assimilation results of both FNP and ConvCNP. However, the increased quantity of observations and information will bring the benefits, resulting in improved performance for some variables in out-of-domain settings, although the number of such variables decreases as the resolution increases. FNP consistently outperforms ConvCNP in terms of the number of variables showing improvement and even exhibits superior performance to fine-tuned versions of other models in some variables. \textit{This demonstrates the excellent out-of-domain generalization capability of our method in adapting to changes in resolution}, and this capability is also applicable to changes in background resolution theoretically.

Figure~\ref{fig:visual_q700_721} presents the visualization of assimilation results by different models for q700, with the visualization date-time randomly selected at 2018-04-02 06:00 UTC. The first row displays the ERA5 (ground truth), background, background error (background minus ERA5), and raw observations with a resolution of 0.25°. Other rows show the analysis, analysis increment (analysis minus background), and analysis error (analysis minus ERA5) obtained through data assimilation by different models, as well as the interpolated observations for Adas and analysis variances for ConvCNP and FNP. The background is smoother compared with ERA5, and the background error shows high spatial variability. It can be observed that \textit{FNP accurately captures the distribution pattern of the background error, leading to analysis with rich high-frequency information and significantly reduced analysis error}. The comparison with ConvCNP, which also assimilates raw observations directly, confirms that FNP's excellent ability to capture high-frequency features does not sorely rely on higher-resolution observations. Furthermore, the smaller analysis variance also indicates a lower uncertainty in state estimation achieved by FNP. More visualizations with different variables and resolutions are shown in Appendix~\ref{appendix: visualization} 

\subsection{Generalization to Observational Information Reconstruction}

Theoretically, the functional representation learned based on observational conditions can be directly decoded through MLPs and output reconstruction results for the observational information without fine-tuning. Therefore, we evaluated the reconstruction performance of different models in the absence of the background conditions, as shown in Table~\ref{reconstruction}. Similarly, Adas pre-trained on data assimilation task, cannot be directly used for information reconstruction. Hence, the table only presents the performance of retrained Adas, while both FNP and ConvCNP show results with and without fine-tuning. \textit{The fine-tuned FNP achieves SOTA performance across all metrics, while FNP without fine-tuning also demonstrates good generalization.}

\begin{table}[h]
  \caption{Performance comparison and ablation study in information reconstruction with different observation ratios.}
  \label{reconstruction}
  \centering
  \resizebox{\linewidth}{!}{
  \setlength{\extrarowheight}{3pt}
  \begin{tabular}{l|c|cc|cccccccc}
    \toprule
    \multirow{2}{*}{Model} & \multirow{2}{*}{Ratio} & \multirow{2}{*}{MSE} & \multirow{2}{*}{MAE} & \multicolumn{8}{c}{RMSE} \\
    & & & & z500 & t850 & t2m & u10 & v10 & u500 & v500 & q700 ($10^{-4}$) \\
    \midrule
    Adas~\cite{chen2023towards} & 10\% & 0.0370 & 0.1006 & 46.592 & 1.0482 & 1.3715 & \underline{1.1539} & \underline{1.1711} & 2.0015 & 1.9932 & 9.5399 \\
    ConvCNP~\cite{gordon2019convolutional} w/o fine-tuning & 10\% & 0.0376 & 0.1042 & 67.098 & 1.0630 & 1.4864 & 1.2482 & 1.2756 & 1.9896 & 2.0312 & 9.3894 \\
    ConvCNP~\cite{gordon2019convolutional} & 10\% & \underline{0.0350} & \underline{0.0933} & 59.864 & \underline{0.8789} & \underline{0.8219} & 1.2165 & 1.2145 & 1.9715 & 2.0069 & \underline{6.8643} \\
    FNP (ours) w/o fine-tuning & 10\% & 0.0355 & 0.1014 & \underline{45.835} & 1.0378 & 1.3382 & 1.1723 & 1.2024 & \underline{1.9559} & \underline{1.9075} & 9.1517 \\
    FNP (ours) & 10\% & \textbf{0.0291} & \textbf{0.0876} & \textbf{35.491} & \textbf{0.7298} & \textbf{0.8027} & \textbf{0.9411} & \textbf{0.9716} & \textbf{1.7464} & \textbf{1.7564} & \textbf{6.7258} \\
    \bottomrule
  \end{tabular}
  }
\end{table}

\subsection{Ablation Study}

We conduct ablation experiments on both the designed modules employed in FNP and experimental settings. Table~\ref{ablation study for modules} presents the quantitative performance comparison of FNP when different components are replaced. The overall MSE, MAE, and RMSE metrics of all variables exhibit varying degrees of performance degradation when a specific module in FNP is replaced. FNP achieves the best performance when these designed components work in synergy and mutually reinforce each other.

\begin{table}[h]
  \caption{Ablation study of different components in FNP for data assimilation with 1.40625° resolution.}
  \label{ablation study for modules}
  \centering
  \resizebox{\linewidth}{!}{
  \setlength{\extrarowheight}{3pt}
  \begin{tabular}{l|c|cc|cccccccc}
    \toprule
    \multirow{2}{*}{Model} & \multirow{2}{*}{Resolution} & \multirow{2}{*}{MSE} & \multirow{2}{*}{MAE} & \multicolumn{8}{c}{RMSE} \\
    & & & & z500 & t850 & t2m & u10 & v10 & u500 & v500 & q700 ($10^{-4}$) \\
    \midrule
    Background & 1.40625° & 0.0288 & 0.0861 & 45.455 & 0.7200 & 0.7790 & 0.9336 & 0.9645 & 1.7278 & 1.7535 & 6.7220 \\
    \midrule
    FNP w/o NFL & 1.40625° & 0.0230 & 0.0749 & 30.040 & 0.6733 & 0.7504 & 0.8212 & 0.8254 & 1.5459 & 1.5566 & 6.4941 \\
    FNP w/o DAM & 1.40625° & 0.0214 & 0.0702 & 30.189 & 0.6816 & 0.7138 & 0.8248 & 0.8090 & 1.4674 & 1.4831 & 6.5631 \\
    FNP w/o SVD & 1.40625° & 0.0239 & 0.0757 & 27.588 & 0.6683 & 0.7555 & 0.7805 & 0.8430 & 1.5743 & 1.5789 & 6.5538 \\
    FNP & 1.40625° & \textbf{0.0202} & \textbf{0.0664} & \textbf{26.910} & \textbf{0.6600} & \textbf{0.7049} & \textbf{0.7558} & \textbf{0.7665} & \textbf{1.4263} & \textbf{1.4450} & \textbf{6.4691} \\
    \bottomrule
  \end{tabular}
  }
\end{table}

Ablation study on the experimental settings is conducted by changing the forecast lead time of the background and the ratio of observations while keeping a fixed resolution of 1.40625°. Table~\ref{ablation study for experimental settings} provides a quantitative performance comparison of different models when the observation proportion is reduced to 1\% and when the forecast lead time of the background is extended to 48 hours. In these scenarios, all the models exhibit robustness and consistently improve the background. When the number of observations decreases or the background error increases, the amount of conditional information they can provide becomes less, so it is reasonable to observe an increase in analysis error compared to Table~\ref{arbitrary data assimilation}. Similarly, \textit{FNP achieves SOTA performance in terms of overall MSE, MAE, and RMSE for the majority of variables}.

\begin{table}[h]
  \caption{Ablation study on the forecast lead time of the background and the ratio of observations.}
  \label{ablation study for experimental settings}
  \centering
  \resizebox{\linewidth}{!}{
  \setlength{\extrarowheight}{3pt}
  \begin{tabular}{l|c|cc|cccccccc}
    \toprule
    \multirow{2}{*}{Model} & \multirow{2}{*}{Ratio} & \multirow{2}{*}{MSE} & \multirow{2}{*}{MAE} & \multicolumn{8}{c}{RMSE} \\
    & & & & z500 & t850 & t2m & u10 & v10 & u500 & v500 & q700 ($10^{-4}$) \\
    \midrule
    24h Background & - & 0.0288 & 0.0861 & 45.455 & 0.7200 & 0.7790 & 0.9336 & 0.9645 & 1.7278 & 1.7535 & 6.7220 \\
    \midrule
    Adas~\cite{chen2023towards} & 1\% & 0.0272 & 0.0835 & \textbf{30.595} & 0.7178 & 0.7771 & 0.9115 & 0.9406 & 1.6881 & 1.7141 & \textbf{6.7074} \\
    ConvCNP~\cite{gordon2019convolutional} & 1\% & 0.0283 & 0.0856 & 43.272 & 0.7186 & 0.7785 & 0.9286 & 0.9574 & 1.7203 & 1.7428 & 6.7173 \\
    FNP (ours) & 1\% & \textbf{0.0269} & \textbf{0.0812} & 30.678 & \textbf{0.6971} & \textbf{0.7708} & \textbf{0.9069} & \textbf{0.9351} & \textbf{1.6586} & \textbf{1.6778} & 6.7084 \\
    \midrule
    48h Background & - & 0.0467 & 0.1148 & 85.017 & 0.9103 & 0.9283 & 1.2525 & 1.2882 & 2.3583 & 2.3856 & 8.2664 \\
    \midrule
    Adas~\cite{chen2023towards} & 10\% & 0.0305 & 0.0840 & 30.473 & \textbf{0.7355} & 0.8134 & 1.0047 & 1.0176 & 1.7747 & 1.7217 & 7.8209 \\
    ConvCNP~\cite{gordon2019convolutional} & 10\% & 0.0334 & 0.0918 & 40.378 & 0.7774 & 0.8619 & 0.9251 & 0.9422 & 1.8112 & 1.8160 & 7.7774 \\
    FNP (ours) & 10\% & \textbf{0.0252} & \textbf{0.0743} & \textbf{28.883} & 0.7459 & \textbf{0.7810} & \textbf{0.8307} & \textbf{0.8517} & \textbf{1.6285} & \textbf{1.6348} & \textbf{7.6078} \\
    \bottomrule
  \end{tabular}
  }
\end{table}

\section{Conclusions}
\label{section: conclusions}

In summary, we present FNP that can assimilate observations with arbitry resolution. The outstanding performance and out-of-domain generalization of FNP in data assimilation and observational information reconstruction demonstrate its significant potential and broad application prospects. It not only contributes to the field of data assimilation but also makes meaningful explorations for AI-based end-to-end weather forecasting systems. Our work has certain limitations. Firstly, the observational data used in our experiments are generated through simulations rather than real-world observations. This may lead to differences in model performance when applied in actual scenarios, thus discounting its value for practical application. In fact, due to the complex and diverse nature of real observational data, the data assimilation community lacks relevant benchmarks and large-scale datasets. The establishment of such benchmarks and datasets would be a highly meaningful endeavor, enabling fair comparisons among different models and fostering rapid advancements in the field. Secondly, FNP inherently performs 3D data assimilation without the temporal dimension. Considering the flexibility of the FNP architecture, incorporating the temporal dimension is not challenging and is expected to produce additional benefits. In the future, we will further explore the broader possibilities of data assimilation and end-to-end weather forecasting.

\begin{ack}
This work is supported by National Natural Science Foundation of China (No. 62071127 and 62101137), National Key Research and Development Program of China (No. 2022ZD0160100), Shanghai Natural Science Foundation (No. 23ZR1402900) and Shanghai Municipal Science and Technology Major Project (No.2021SHZDZX0103).
\end{ack}

\bibliographystyle{plain}
\bibliography{references}


\appendix

\newpage
\section{Code Availability}
\label{appendix: code}
The source code is available at \url{https://anonymous.4open.science/r/FNP-10406/}.

\section{More visualizations for data assimilation}
\label{appendix: visualization}

\begin{figure}[h]
    \centering
    \includegraphics[width=\linewidth]{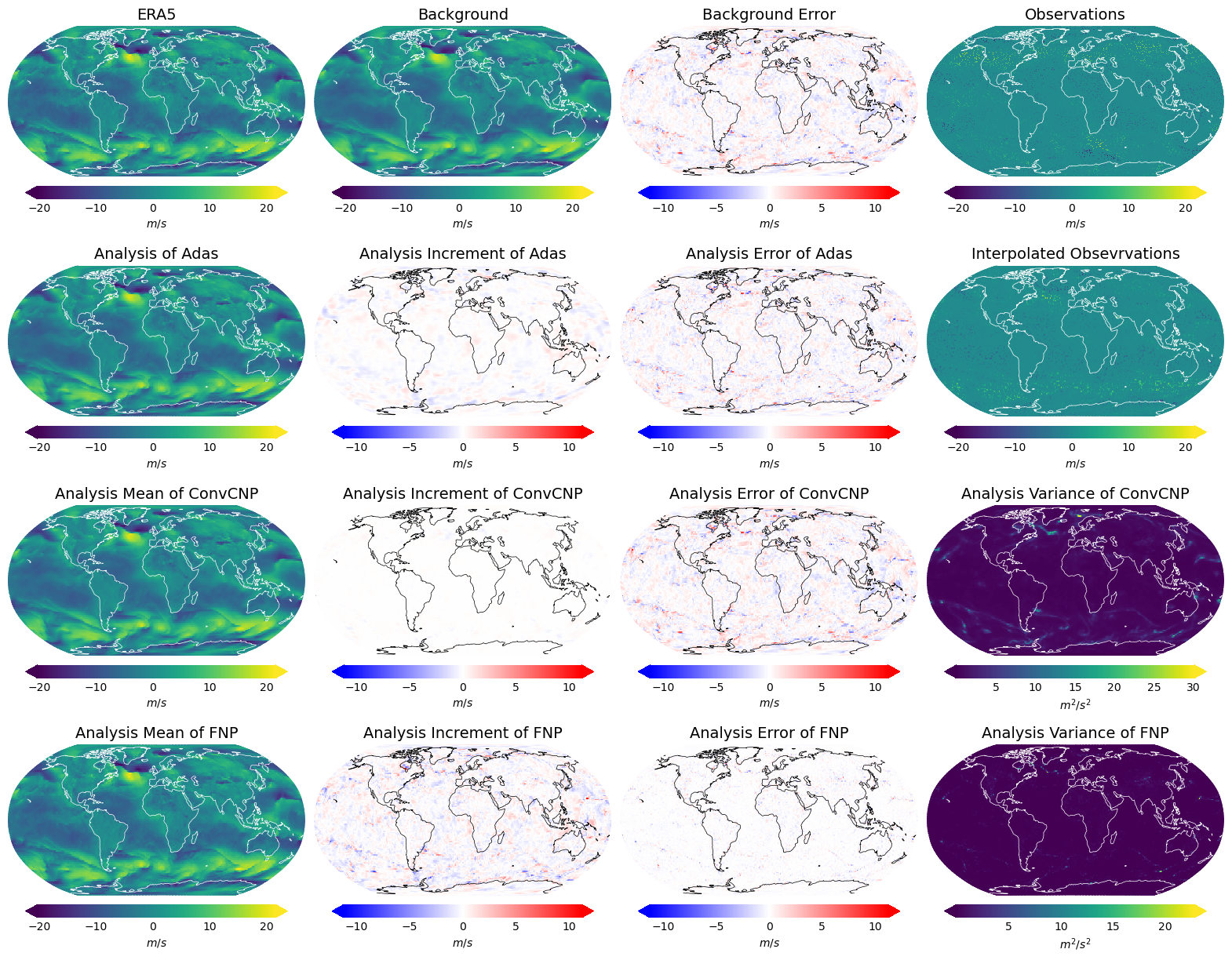}
    \caption{Visualization of assimilation results with 0.25° resolution for u10.}
\end{figure}

\newpage
\begin{figure}[h]
    \centering
    \includegraphics[width=\linewidth]{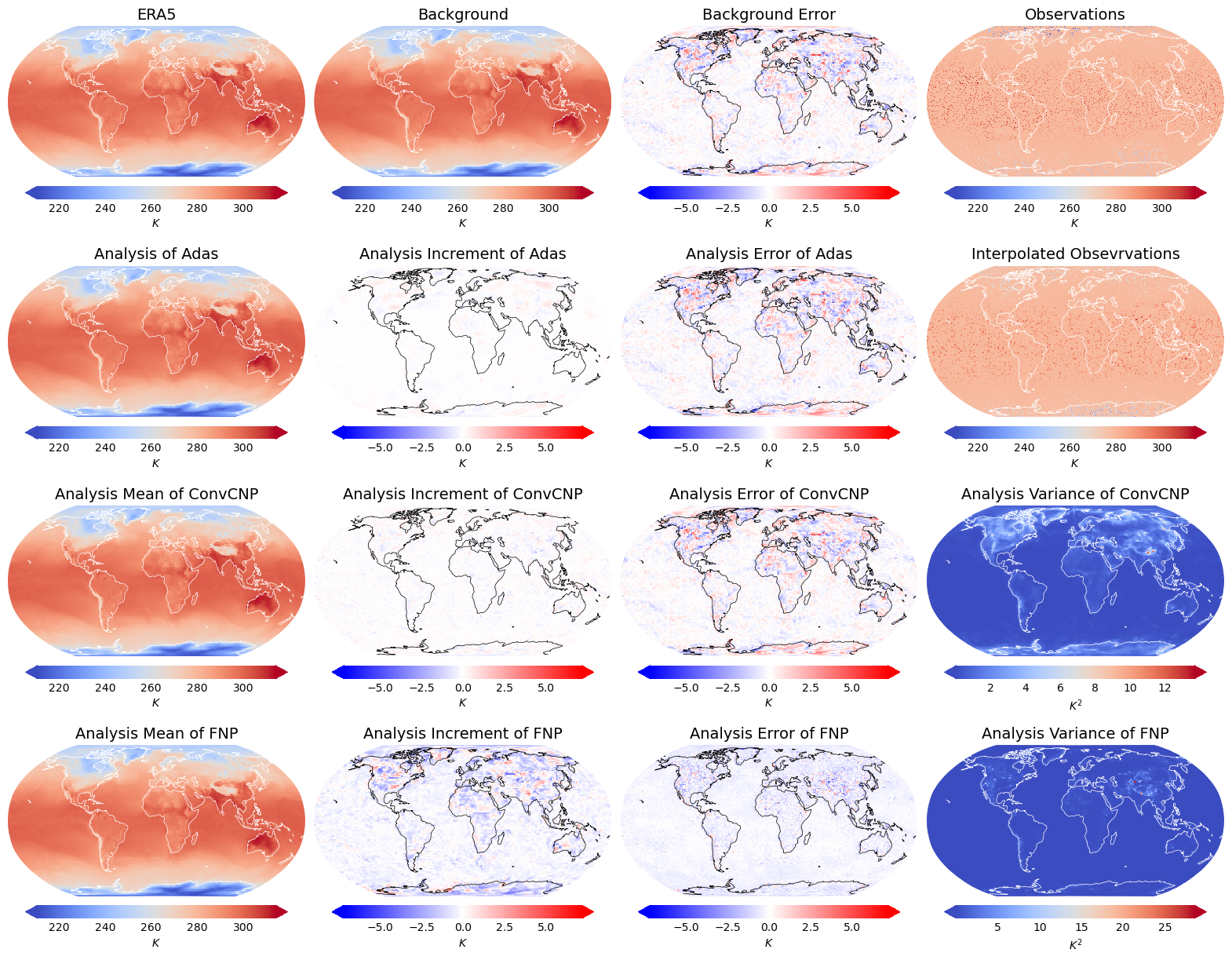}
    \caption{Visualization of assimilation results with 0.25° resolution for t2m.}
\end{figure}

\newpage
\begin{figure}[h]
    \centering
    \includegraphics[width=\linewidth]{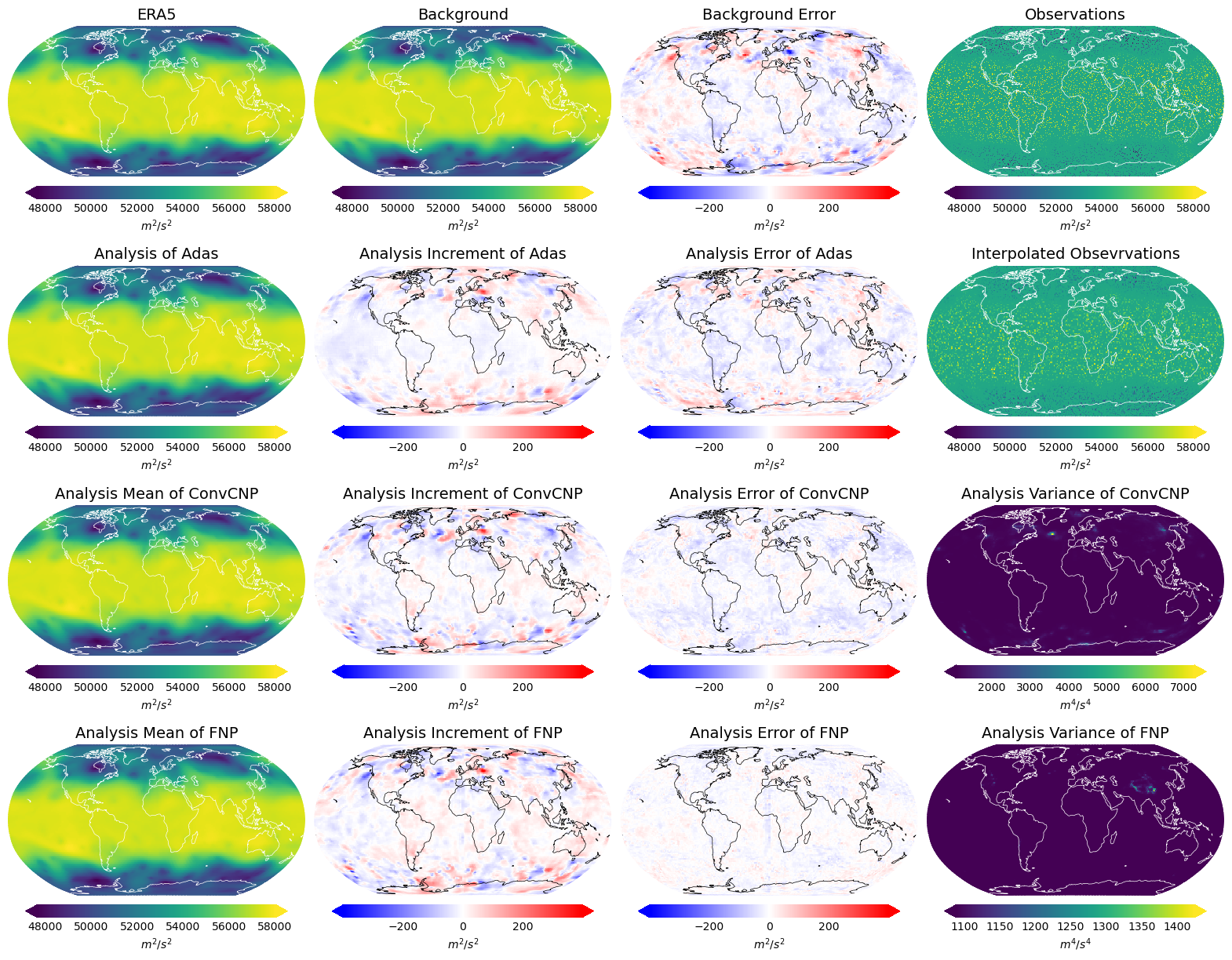}
    \caption{Visualization of assimilation results with 0.25° resolution for z500.}
\end{figure}

\newpage
\begin{figure}[h]
    \centering
    \includegraphics[width=\linewidth]{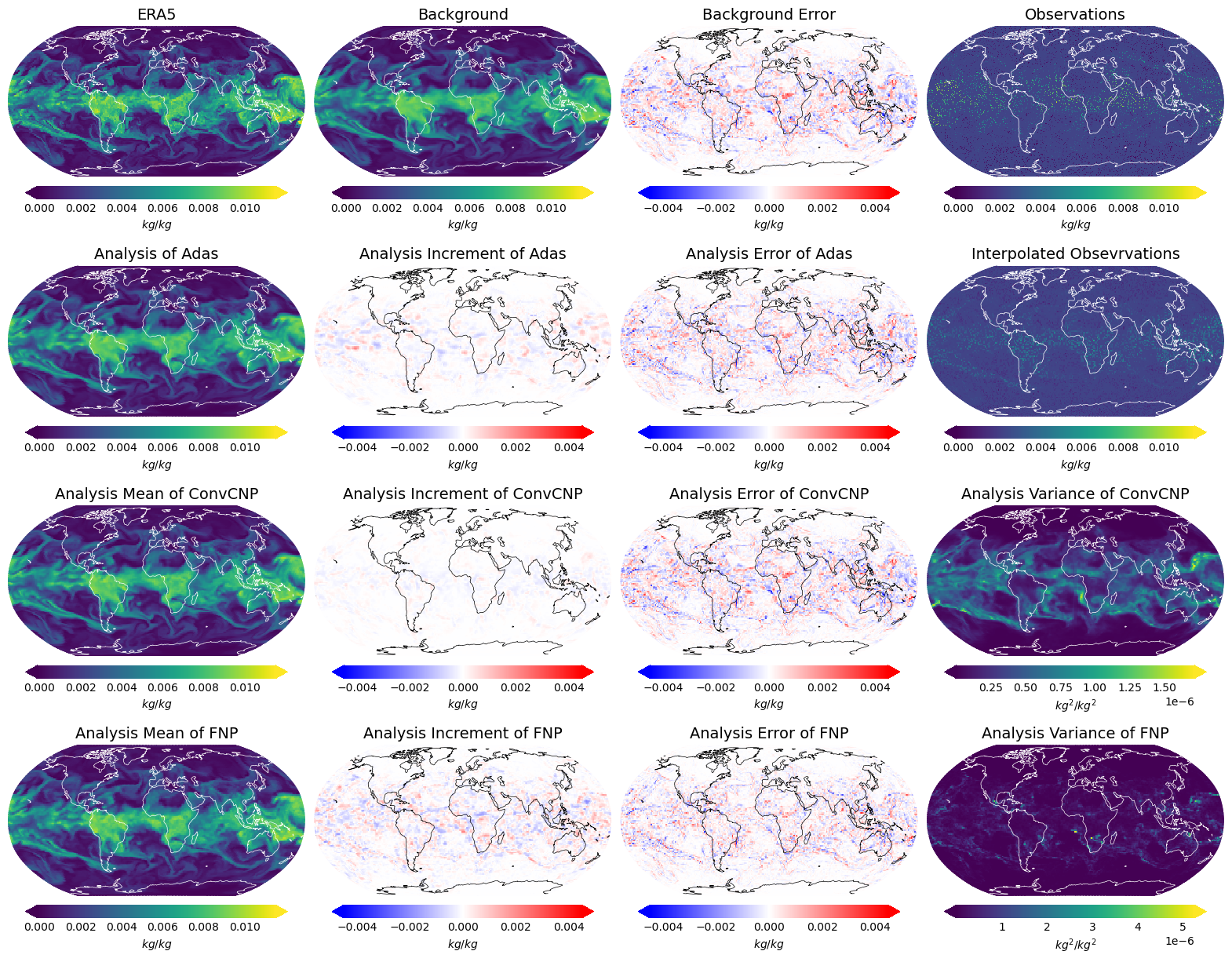}
    \caption{Visualization of assimilation results with 0.703125° resolution for q700.}
\end{figure}

\newpage
\begin{figure}[h]
    \centering
    \includegraphics[width=\linewidth]{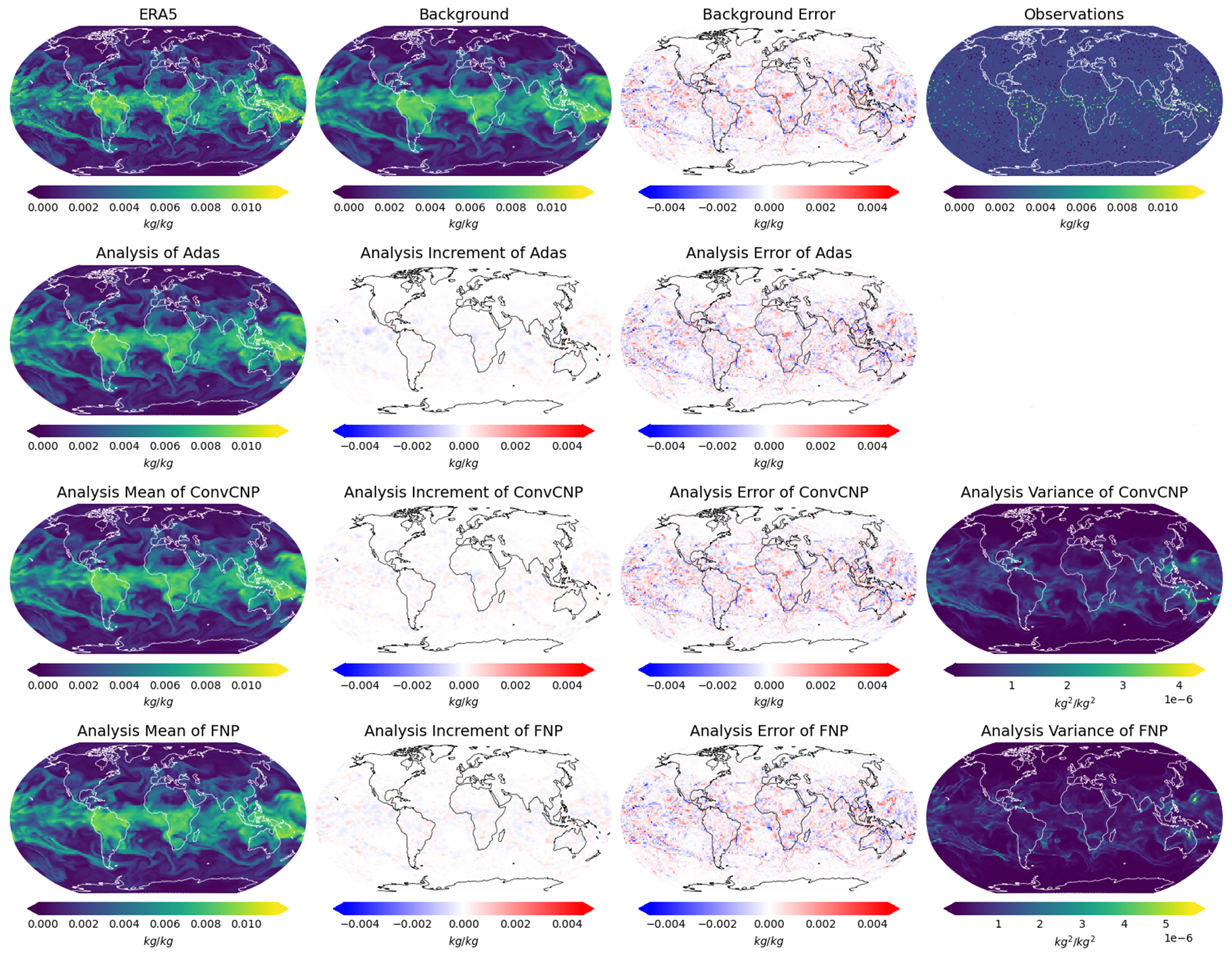}
    \caption{Visualization of assimilation results with 1.40625° resolution for q700.}
\end{figure}

\end{document}